\documentclass[a4paper,fleqn]{cas-sc}

\usepackage[authoryear,longnamesfirst]{natbib}

\def\tsc#1{\csdef{#1}{\textsc{\lowercase{#1}}\xspace}}
\tsc{WGM}
\tsc{QE}

\begin{document}
\let\WriteBookmarks\relax
\def\floatpagepagefraction{1}
\def\textpagefraction{.001}

\shorttitle{Continuous Temporal Energy Representations}

\shortauthors{M. Bengtsson}

\title[mode=title]{Continuous Temporal Energy Representations for Event-Based Signals}

\author[1]{Magnus Bengtsson}

\cormark[1]

\ead{magnus.bengtsson@hb.se}

\credit{Conceptualization, Methodology, Software, Writing}

\affiliation[1]{
    organization={Department of Engineering, University of Borås},
    city={Borås},
    country={Sweden}
}

\cortext[1]{Corresponding author}
\fntext[1]{0000-0002-3283-067X}


\begin{abstract}
Spatio-temporal signals arising from event-driven biological processes, such as surface electromyography (sEMG), exhibit asynchronous activation patterns that are challenging to model using conventional discrete or purely real-valued temporal representations.

In this work, we propose the Continuous Temporal Energy Network (CTEN), a continuous temporal modeling framework for event-driven signals based on latent temporal energy representations. The proposed approach maps asynchronous temporal activity into a continuous latent space, where temporal structure is represented through phase-modulated activations and energy-based temporal projections over finite observation windows.

By transforming sparse event activity into structured continuous temporal energy patterns, the model enables temporal feature extraction without explicit recurrence, discrete spike simulation, or surrogate gradient approximations. The resulting representation remains fully differentiable and compatible with conventional gradient-based optimization.

Experimental evaluation on a synthetic interaural phase difference (IPD) classification task demonstrates stable and competitive performance compared to both spike-based and conventional temporal neural architectures. Additional experiments on the Ninapro sEMG dataset further indicate that the proposed representation generalizes beyond synthetic benchmarks to real-world multichannel biosignals using raw temporal input signals with minimal preprocessing.

Ablation experiments further suggest that the primary representational strength of the framework originates from the continuous temporal energy representation itself, highlighting the effectiveness of differentiable continuous temporal embeddings for event-driven learning tasks. 
\end{abstract}


\begin{highlights}
\item Continuous temporal energy representations for asynchronous event-driven signals
\item Fully differentiable temporal modeling without surrogate gradient approximations
\item Phase-modulated latent embeddings with energy-based temporal aggregation
\item Competitive performance on synthetic IPD classification and real-world sEMG signals
\item Reduced training time compared to spiking neural network baselines
\end{highlights}



\maketitle





\clearpage 




\section{Introduction}

Neural activity is often modeled at the level of discrete spike events, as in spiking neural networks (SNNs) \cite{gerstner2014neuronal,neftci2019surrogate}, where information is represented through temporally precise firing patterns. Such models aim to capture the dynamics of individual neurons and their interactions, and have shown promise as a biologically inspired framework for event-driven computation. 

However, in many practical settings, neural signals are not observed at this microscopic level. Instead, measurements such as electromyography (EMG) \cite{konrad2005abc} or surface electromyography (sEMG) \cite{deLuca2002surface} reflect aggregate activity arising from large populations of neurons, resulting in continuous signals with structured temporal patterns.

This distinction motivates an alternative perspective. Rather than explicitly modeling spike generation, it may be sufficient to represent neural activity at the level of observable macroscopic signals. In this setting, information is not necessarily encoded in individual spike events, but in the temporal distribution of activation over finite time intervals.

From this viewpoint, SNNs can be seen as one possible approach for modeling event-driven signals, but they do not directly address the problem of representing the probability of event occurrence in continuous time. In many applications, including biosignal analysis and control of biomechanical systems, it is more relevant to estimate when meaningful events are likely to occur, rather than to reconstruct the exact underlying spike sequence. This suggests the need for complementary modeling frameworks that operate on a continuous temporal representation.

A key advantage of continuous representations is that they are naturally compatible with standard optimization techniques. Unlike discrete spike-based models, which often require surrogate gradient methods or specialized training procedures, continuous formulations allow the use of conventional gradient-based learning and efficient GPU acceleration.

In this work, we propose a continuous wave-based representation for modeling asynchronous neural activity. The model encodes input signals as wave-based fields, where phase-modulated continuous wave representations capture temporal structure within finite temporal regions. Through phase-modulated latent interactions and energy-based projection, the model induces structured activation patterns that reflect both temporal localization and relational dependencies.

The key contribution of this work is the introduction of a continuous temporal energy representation that enables differentiable modeling of asynchronous event-driven signals without explicit recurrence or surrogate gradient approximations.

The proposed framework is particularly motivated by event-driven biosignals such as sEMG, where the objective is to extract robust temporal features for downstream control tasks, including the operation of biomechanical prostheses and exoskeleton systems.

The objective is not to replicate biological mechanisms at the level of individual neurons, but to capture the functional temporal structure of event-driven signals in a continuous and computationally efficient manner.

Unlike standard static embeddings, the proposed formulation preserves temporal structure through continuous latent energy representations over finite observation windows.

The present study focuses on a controlled synthetic benchmark in order to isolate the properties of the proposed temporal representation independently of sensor-specific artifacts and dataset-dependent variability.

\subsection{Interaural Phase Difference (IPD) Classification Task}

The experimental setup is conceptually inspired by educational and research-oriented examples of temporal coding and binaural processing presented in the COSYNE 2022 tutorial on spiking neural networks \cite{goodman2022cosyne}. In these examples, interaural phase differences (IPD) are encoded through temporally structured neural activity and processed using spiking network models.

In contrast to the biologically grounded formulations used in the tutorial, the present work adopts a simplified and discretized classification setting, where IPD is mapped to discrete output classes. This abstraction allows controlled evaluation of temporal representation mechanisms while preserving the essential phase-based structure of the input signals.

The proposed models are evaluated on a synthetic IPD classification task, which serves as a controlled benchmark for event-based temporal processing. The setup isolates the model's ability to capture temporal structure independently of biological constraints by removing confounding biological or environmental factors, while retaining the core phase-based encoding principle.

In this setup, each sample consists of a temporal event tensor
\[
x \in \mathbb{R}^{T \times D},
\]
where $T$ denotes the number of time steps and $D = 2 \times N_{\text{ear}}$ corresponds to the number of input channels. The channels are divided into two groups representing left and right ``ears'', each containing $N_{\text{ear}} = 200$ channels, resulting in $D = 400$ input channels.

Spike trains are generated using phase-shifted sinusoidal processes, where a global interaural phase difference parameter $\Delta \phi$ controls the relative delay between the two channel groups. This parameter is sampled uniformly and discretized into $C = 12$ classes, forming the target labels for classification.

The temporal resolution is fixed to $T = 100$ time steps over a duration of $0.1$ seconds, corresponding to a discretization step of $\Delta t = 10^{-3}$ seconds.

\paragraph{Model configurations.}
For the proposed Continuous Temporal Energy Network (CTEN), the input is projected into a latent space of dimension $H = 160$, followed by a low-rank interference module with rank $R = 48$. The resulting representation is processed using temporal aggregation and a lightweight multilayer perceptron for classification.

For comparison, the spiking neural network (SNN) baseline employs a single hidden spiking layer with $H = 400$ neurons, followed by a linear readout layer. Temporal dynamics are modeled through leaky membrane integration with surrogate gradient-based training.

This experimental setup enables a direct comparison between continuous wave-based representations and discrete spike-based processing under identical input conditions, highlighting differences in how temporal information is encoded and utilized.
\section{Theory}

\subsection{Asynchronous Spike Representation}

Let a spike train be defined as a set of discrete events:
\begin{equation}
S = \{t_i\}_{i=1}^N,
\end{equation}
where $t_i$ denotes the time of the $i$-th spike.

A spike train can equivalently be represented as a sum of Dirac impulses:
\begin{equation}
x(t) = \sum_i \delta(t - t_i),
\end{equation}
which is a standard formulation in signal processing and computational neuroscience \cite{gerstner2014neuronal}.

In conventional SNNs, information is encoded directly in these event times. However, this representation is inherently sparse and non-differentiable.

\subsection{Continuous Temporal Embedding}

This formulation serves as a conceptual bridge between discrete spike representations and continuous temporal embeddings. 

We consider a continuous representation $\psi(t)$ of the spike train via convolution with a smoothing kernel $k(t)$:
\begin{equation}
\psi(t) = \sum_i k(t - t_i).
\end{equation}

This representation is not explicitly used in the proposed model, but serves to motivate the transition from discrete spike events to continuous wave-based representations. 

The formulation corresponds to a kernel smoothing operation commonly used in spike train analysis and density estimation \cite{shimazaki2010kernel}.

A common choice is the Gaussian kernel:
\begin{equation}
k(t) = \exp\left(-\frac{t^2}{2\sigma_k^2}\right),
\end{equation}
which yields a smooth temporal activation profile.

This transformation converts discrete events into a continuous signal while preserving temporal locality and approximate event timing.
Gaussian kernels are widely used due to their optimal smoothing properties and connection to radial basis functions \cite{bishop2006pattern}.
Importantly, the proposed model does not rely on explicit kernel convolution; 
instead, it learns a continuous representation through phase-modulated wave dynamics.
\subsection{Energy-Based Interpretation}

We interpret the squared amplitude of the signal as an energy density:
\begin{equation}
\rho(t) = \psi(t)^2.
\end{equation}

This allows us to define global temporal descriptors:

\paragraph{Energy}
\begin{equation}
E = \int \psi(t)^2\, dt,
\end{equation}
This corresponds to the signal energy, i.e., the squared $L^2$ norm, a fundamental quantity in signal processing \cite{oppenheim1999signals}.

\paragraph{Temporal Mean}
\begin{equation}
\mu = \frac{1}{E} \int t \psi(t)^2\, dt,
\end{equation}

\paragraph{Temporal Variance}
\begin{equation}
\sigma^2 = \frac{1}{E} \int (t - \mu)^2 \psi(t)^2\, dt.
\end{equation}

These quantities correspond to the first and second moments of a probability density function \cite{papoulis2002probability}.

\subsection{Interpretation as Asynchronous Encoding}

The continuous signal $\psi(t)$ encodes spike activity such that:

\begin{itemize}
\item $\mu$ captures the average activation time,
\item $\sigma$ captures temporal spread (asynchrony),
\item $E$ captures activation strength.
\end{itemize}

Thus, asynchronous spike timing is embedded into a continuous temporal structure.

\subsection{Limitations of Purely Real-Valued Representations}

The continuous representation $\psi(t)$ defined above captures temporal structure through smoothing and energy-based statistics. However, this formulation is inherently real-valued and therefore loses phase information.

As a consequence, independent components of the signal cannot interact in a structured manner. In particular, the squared magnitude
\begin{equation}
    \rho(t) = \psi(t)^2
\end{equation}
does not contain cross-terms between different components, limiting the representational capacity.

This motivates an extension toward a complex-valued formulation, where phase interactions enable interference effects between latent components.

\section{Continuous Temporal Energy Network}

To address the limitations of purely real-valued temporal embeddings, we extend the representation to a complex-valued wave formulation. This enables phase-dependent latent representations and nonlinear interactions between temporal components within a continuous energy-based framework.

\subsection{Input Representation}

Let the input be a temporal signal
\begin{equation}
    x(t) \in \mathbb{R}^C, \quad t \in [0,T]
\end{equation}
where $C$ denotes the number of input channels.

\subsection{Linear Projection}

The input is projected into a latent space of dimension $H$:
\begin{equation}
    h(t) = W^\top x(t), \quad W \in \mathbb{R}^{C \times H}
\end{equation}

\subsection{Phase Representation}

Each latent unit is assigned a phase:
\begin{equation}
    \theta_h(t) = \omega_h t + \phi_h
\end{equation}
where $\omega_h$ is the angular frequency and $\phi_h$ the phase offset.

\subsection{Complex Wave Field}

Complex-valued representations are widely used in signal processing and wave physics, where amplitude and phase jointly encode information \cite{mallat1999wavelet}. This allows interactions between latent components through phase alignment and interference.

The latent representation is constructed as a complex-valued wave field:
\begin{equation}
    \psi_h(t) = \tilde{h}_h(t)\, e^{i \theta_h(t)}
\end{equation}
where
\begin{equation}
    \tilde{h}_h(t) = \tanh\!\left(h_h(t)\right)
\end{equation}
denotes a bounded activation of the projected signal, and
\begin{equation}
    \theta_h(t) = \omega_h t + \phi_h
\end{equation}
is a learnable phase function with angular frequency $\omega_h$ and phase offset $\phi_h$.

In real-valued form, the complex wave field can be expressed as:
\begin{align}
    \psi_h^{(R)}(t) &= \tilde{h}_h(t)\cos(\theta_h(t)) \\
    \psi_h^{(I)}(t) &= \tilde{h}_h(t)\sin(\theta_h(t))
\end{align}

This formulation can be interpreted as a phase-modulated embedding of the latent signal, where temporal structure is encoded through oscillatory components rather than explicit temporal recurrence.
\subsection{Phase-Modulated Latent Interaction}

Interaction between latent components is introduced via a low-rank nonlinear mixing:
\begin{align}
\psi^{(R)}(t) &\leftarrow \psi^{(R)}(t) + \alpha\, \tanh\!\big(\psi^{(R)}(t) W_{\text{int1}} W_{\text{int2}}\big) \\
\psi^{(I)}(t) &\leftarrow \psi^{(I)}(t) + \alpha\, \tanh\!\big(\psi^{(I)}(t) W_{\text{int1}} W_{\text{int2}}\big)
\end{align}
where $\psi(t)$ is treated as a row vector, i.e., $\psi(t) \in \mathbb{R}^{1 \times H}$, and
\begin{equation}
W_{\text{int1}} \in \mathbb{R}^{H \times R}, \quad
W_{\text{int2}} \in \mathbb{R}^{R \times H}
\end{equation}
form a low-rank factorization of the interaction operator.

\subsection{Probability Density and Latent Interaction}

The squared magnitude corresponds to signal energy or intensity, analogous to energy densities in wave-based representations \cite{born1999principles}.

For each latent component, we define the energy representation as
\begin{equation}
P_h(t) = |\psi_h(t)|^2
\end{equation}

The latent energy vector is then given by
\begin{equation}
\mathbf{P}(t) = \big(P_1(t), \dots, P_H(t)\big)
\end{equation}

The vector $\mathbf{P}(t)$ can be interpreted as a continuous latent energy representation over time.

In the present implementation, the energy projection is computed independently for each latent component. Consequently, the model does not explicitly compute interference terms arising from the superposition of multiple complex-valued components.

However, if the latent representation is interpreted as a combined wave field,
the total energy would take the form
\begin{equation}
|\psi(t)|^2 =
\sum_{h=1}^{H} |\psi_h(t)|^2
+
\sum_{h \neq k}
\Re\!\left(\psi_h(t)\psi_k^*(t)\right)
\end{equation}
where the cross-terms correspond to phase-dependent interactions between latent components \cite{goodman2005introduction}.

Although these explicit interference terms are not directly computed in the current formulation, the phase-modulated latent dynamics and nonlinear mixing layers still enable structured interactions between latent representations prior to energy projection.

The interference expression above should therefore be interpreted primarily as a conceptual motivation for phase-dependent latent interactions rather than as an explicit computational component of the implemented model.

\subsection{Feature Extraction}

From the energy field $\mathbf{P}(t) \in \mathbb{R}^H$, 
we compute global temporal descriptors.

\paragraph{Temporal mean representation}
\begin{equation}
\mathbf{c} =
\frac{1}{T}
\sum_{t=1}^{T}
\mathbf{P}(t)
\end{equation}

\paragraph{Maximum activation}
\begin{equation}
\mathbf{u} =
\max_t \mathbf{P}(t)
\end{equation}

The aggregated feature vector is defined as
\begin{equation}
\mathbf{z} =
\mathrm{concat}(\mathbf{c}, \mathbf{u})
\in \mathbb{R}^{2H}.
\end{equation}

\subsection{Classification}

The final prediction is obtained via a feedforward classifier:
\begin{equation}
\hat{y} = \mathrm{softmax}\!\big(f_{\theta}(\mathbf{z})\big),
\end{equation}
where $f_{\theta}$ denotes a multilayer perceptron (MLP) with learnable parameters $\theta$.

In the present implementation, the classifier corresponds
to a two-layer feedforward network of the form
\begin{equation}
f_{\theta}(\mathbf{z}) = W_2\, \sigma(W_1 \mathbf{z} + b_1) + b_2,
\end{equation}
where $\sigma(\cdot)$ is a nonlinear activation function (ReLU in our case).

\subsection{Model Interpretation}

The model represents each input event as a wave field over time and latent dimensions. 
Temporal structure is encoded through continuous phase-modulated energy representations over finite temporal windows.

Notably, the model does not rely on explicit temporal recurrence or causal state propagation.

To make the proposed formulation explicit, Algorithm~\ref{alg:wavenet} summarizes the full forward computation of the model. The algorithm highlights how discrete spike inputs are transformed into a continuous complex-valued wave representation, followed by interference, energy projection, and feature aggregation.

\begin{figure}[htbp]
\centering
\fbox{
\begin{minipage}{0.95\linewidth}

\textbf{Algorithm 1: Forward pass of the proposed Continuous Temporal Energy Network (CTEN)}

\vspace{0.5em}

\textbf{Input:}

Spike tensor $x \in \mathbb{R}^{B \times T \times D}$

Projection weights $W \in \mathbb{R}^{D \times H}$

Wave parameters $\omega \in \mathbb{R}^{H}$, $\phi \in \mathbb{R}^{H}$

Low-rank interaction matrices
$W_{\text{int1}} \in \mathbb{R}^{H \times R}$,
$W_{\text{int2}} \in \mathbb{R}^{R \times H}$

\vspace{0.5em}

\textbf{Output:}

Predictions $\hat{y} \in \mathbb{R}^{B \times C}$

\vspace{0.5em}

\begin{enumerate}

\item Project input:
\[
h(t) \leftarrow \tanh(x(t)W)
\]

\item Compute phase-modulated embedding:
\[
\psi_{\text{real}}(t)
=
h(t)\odot \cos(\omega t + \phi)
\]

\[
\psi_{\text{imag}}(t)
=
h(t)\odot \sin(\omega t + \phi)
\]

\item Apply latent interaction:
\[
\psi_{\text{real}}
\leftarrow
\psi_{\text{real}}
+
\alpha\tanh(
\psi_{\text{real}}
W_{\text{int1}}
W_{\text{int2}})
\]

\[
\psi_{\text{imag}}
\leftarrow
\psi_{\text{imag}}
+
\alpha\tanh(
\psi_{\text{imag}}
W_{\text{int1}}
W_{\text{int2}})
\]

\item Compute energy representation:
\[
\mathbf{P}(t)
=
\psi_{\text{real}}(t)^2
+
\psi_{\text{imag}}(t)^2
\]

\item Aggregate temporal statistics:
\[
\mathbf{c}
=
\frac{1}{T}
\sum_{t=1}^{T}
\mathbf{P}(t)
\]

\[
\mathbf{u}
=
\max_t \mathbf{P}(t)
\]

\item Concatenate features:
\[
\mathbf{f}
=
\mathrm{concat}(\mathbf{c},\mathbf{u})
\]

\item Predict output:
\[
\hat{y}
=
\mathrm{MLP}(\mathbf{f};\theta)
\]

\end{enumerate}

\end{minipage}
}
\caption{Forward pass of the proposed Continuous Temporal Energy Network (CTEN).}
\label{alg:wavenet}
\end{figure}
\newpage
\section{Results}
\subsection{Ablation Study}

To analyze the contribution of the individual architectural components of the proposed Continuous Temporal Energy Network (CTEN), we conducted a multi-seed ablation study on the synthetic IPD classification task.

The evaluated model variants are summarized in Table~\ref{tab:ablation_components}, while the corresponding quantitative results are presented in Table~\ref{tab:ablation}. The ablations investigate the influence of phase modulation, latent interaction dynamics, and temporal aggregation mechanisms on overall classification performance and training stability.

In particular, we evaluate:
\begin{itemize}
\item removal of the latent interaction module,
\item removal of phase modulation,
\item mean-only temporal aggregation,
\item max-only temporal aggregation.
\end{itemize}

All experiments were evaluated across twenty random seeds in order to assess both predictive performance and convergence stability.

\begin{table}[width=\linewidth,cols=6,pos=h]
\caption{Multi-seed ablation study of CTEN on the synthetic IPD classification task. Results are reported as mean accuracy $\pm$ standard deviation over twenty random seeds.}
\label{tab:ablation}

\begin{tabular*}{\tblwidth}{@{}lccccc@{}}
\toprule
\textbf{Model Variant} &
\textbf{Parameters} &
\textbf{Mean Acc. (\%)} &
\textbf{Std} &
\textbf{Best} &
\textbf{Time (s)} \\
\midrule

CTEN (full model) & 122316 & 91.78 & 1.07 & 93.8 & 4.95 \\
Without interaction & 106956 & 92.11 & 0.89 & 93.8 & 3.22 \\
Without phase modulation & 122316 & 91.47 & 1.42 & 93.2 & 4.14 \\
Mean pooling only & 101836 & 91.90 & 1.43 & 94.6 & 4.76 \\
Max pooling only & 101836 & 91.53 & 0.96 & 93.0 & 4.71 \\

\bottomrule
\end{tabular*}
\end{table}

\begin{table}[width=\linewidth,cols=6,pos=h]
\caption{Overview of the architectural components included in each ablation variant of CTEN.}
\label{tab:ablation_components}

\begin{tabular*}{\tblwidth}{@{}lccccc@{}}
\toprule

\textbf{Model Variant} &
\textbf{Wave Phase} &
\textbf{Interaction} &
\textbf{Energy Projection} &
\textbf{Pooling} &
\textbf{Classifier} \\

\midrule

CTEN (full model)
& Yes
& Yes
& Yes
& Mean + Max
& MLP \\

Without interaction
& Yes
& No
& Yes
& Mean + Max
& MLP \\

Without phase modulation
& No
& Yes
& Yes
& Mean + Max
& MLP \\

Mean pooling only
& Yes
& Yes
& Yes
& Mean
& MLP \\

Max pooling only
& Yes
& Yes
& Yes
& Max
& MLP \\

\bottomrule
\end{tabular*}
\end{table}
The ablation study indicates that the dominant contribution of the proposed framework arises from the continuous temporal energy representation itself. Across multiple random seeds, removing either the latent interaction module or the phase modulation mechanism resulted in only relatively minor performance differences. This suggests that the primary representational capacity originates from the continuous temporal embedding and energy aggregation process rather than from explicit phase-dependent interference effects.

Interestingly, the configuration without the interaction module achieved the highest average accuracy across seeds, while simultaneously reducing the parameter count and computational cost. This observation indicates that the core temporal representation learned by the model is already sufficiently expressive for the present task, and that additional latent interaction dynamics may introduce unnecessary complexity in this setting.

Similarly, the mean-pooling-only and max-pooling-only configurations achieved performance levels comparable to the full model. This suggests that the exact aggregation strategy is of secondary importance compared to the continuous temporal energy formulation itself. The reduced parameter count in these variants originates from the lower-dimensional feature representation provided to the classifier.

The relatively small standard deviations observed across seeds indicate that the proposed framework exhibits stable training dynamics and robust convergence behavior. In contrast, the MLP baseline exhibited near-random classification accuracy despite containing a substantially larger number of parameters. This suggests that static feedforward processing alone is insufficient for effectively capturing the temporal structure of the considered event-based signals.

The poor performance of the MLP baseline further suggests that the task cannot be solved using instantaneous activations alone, but instead requires temporal aggregation of structured event dynamics over finite observation windows.

Overall, the ablation results support the hypothesis that continuous temporal energy representations constitute the primary source of representational power in the proposed framework.

For an interpretable low-dimensional illustration of the proposed transformation,
we refer to Appendix A, where a reduced two-channel configuration is analyzed.

\begin{figure}[h]
    \centering
    \includegraphics[width=\linewidth]{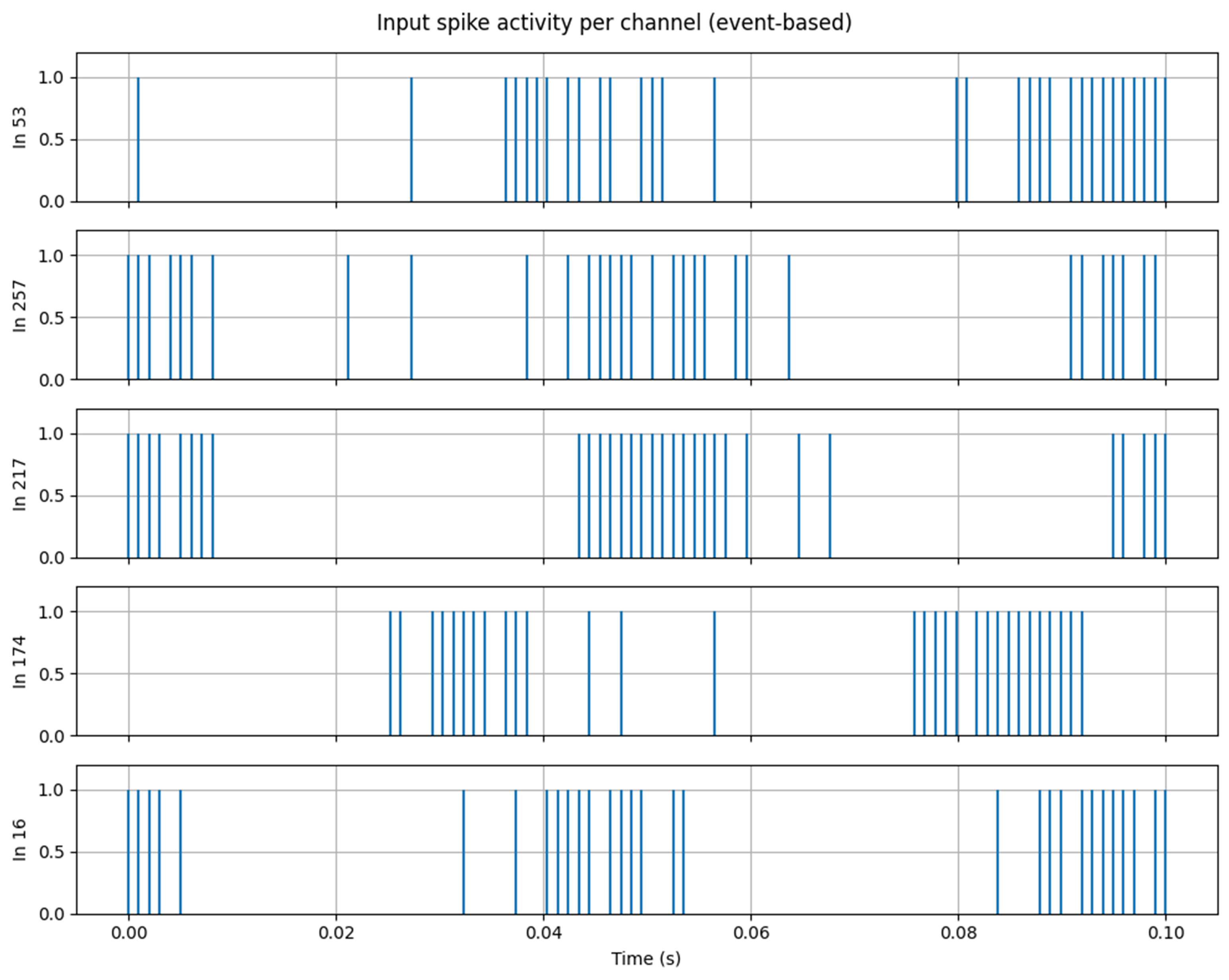}
    \caption{Temporal spike activity for selected input channels.}
    \label{fig:input_spikes}
\end{figure}

To better understand the transformation performed by the proposed model, we first examine the structure of the input spike signals. Figure~\ref{fig:input_spikes} shows the temporal spike activity for a subset of input channels.

As shown in Figure~\ref{fig:input_spikes}, the input consists of sparse and irregular spike events distributed over time. The activity is highly discontinuous, with binary-valued signals indicating the presence or absence of spikes. While some temporal clustering can be observed, the overall representation remains fragmented and lacks an explicit continuous structure.

\begin{figure}[h]
    \centering
    \includegraphics[width=\linewidth]{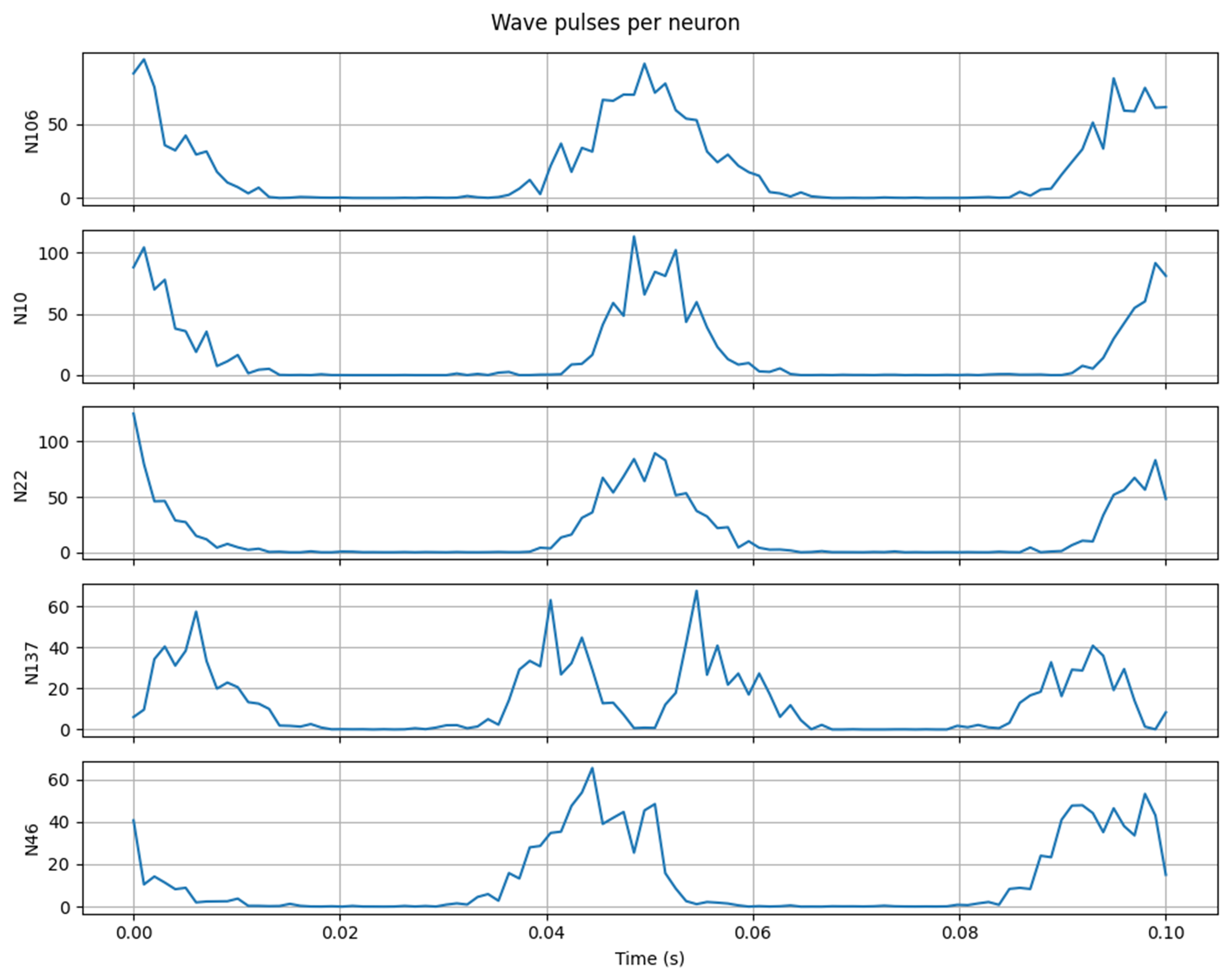}
    \caption{Temporal activation signals $P_h(t) = |\psi_h(t)|^2$ for selected latent neurons.}
    \label{fig:wave_pulses}
\end{figure}

Figure~\ref{fig:wave_pulses} illustrates the corresponding temporal activation profiles produced by the proposed wave-based model. The purpose of this visualization is to highlight how the model transforms discrete spike inputs into a continuous representation.

As shown in Figure~\ref{fig:wave_pulses}, each neuron exhibits localized regions of high activation, forming wave-like pulses over time. These pulses vary in width and amplitude, indicating that the model captures both temporal precision and signal strength. In several neurons, multiple activation peaks are observed, suggesting that the model can represent overlapping or temporally distributed events. Outside the main activation regions, the signal remains close to zero, reflecting a structured and selective representation.

Comparing Figures~\ref{fig:input_spikes} and~\ref{fig:wave_pulses}, it becomes evident that the model transforms sparse, discrete spike trains into smooth, energy-based temporal patterns. This transformation enables a richer representation of temporal structure, where information is encoded in the shape, timing, and intensity of continuous activation signals rather than isolated spike events.

Overall, these results demonstrate that the proposed model provides an effective continuous representation of asynchronous spike activity, preserving temporal information while enabling structured feature extraction in the time domain.

\medskip

\begin{figure}[h]
    \centering
    \includegraphics[width=\linewidth]{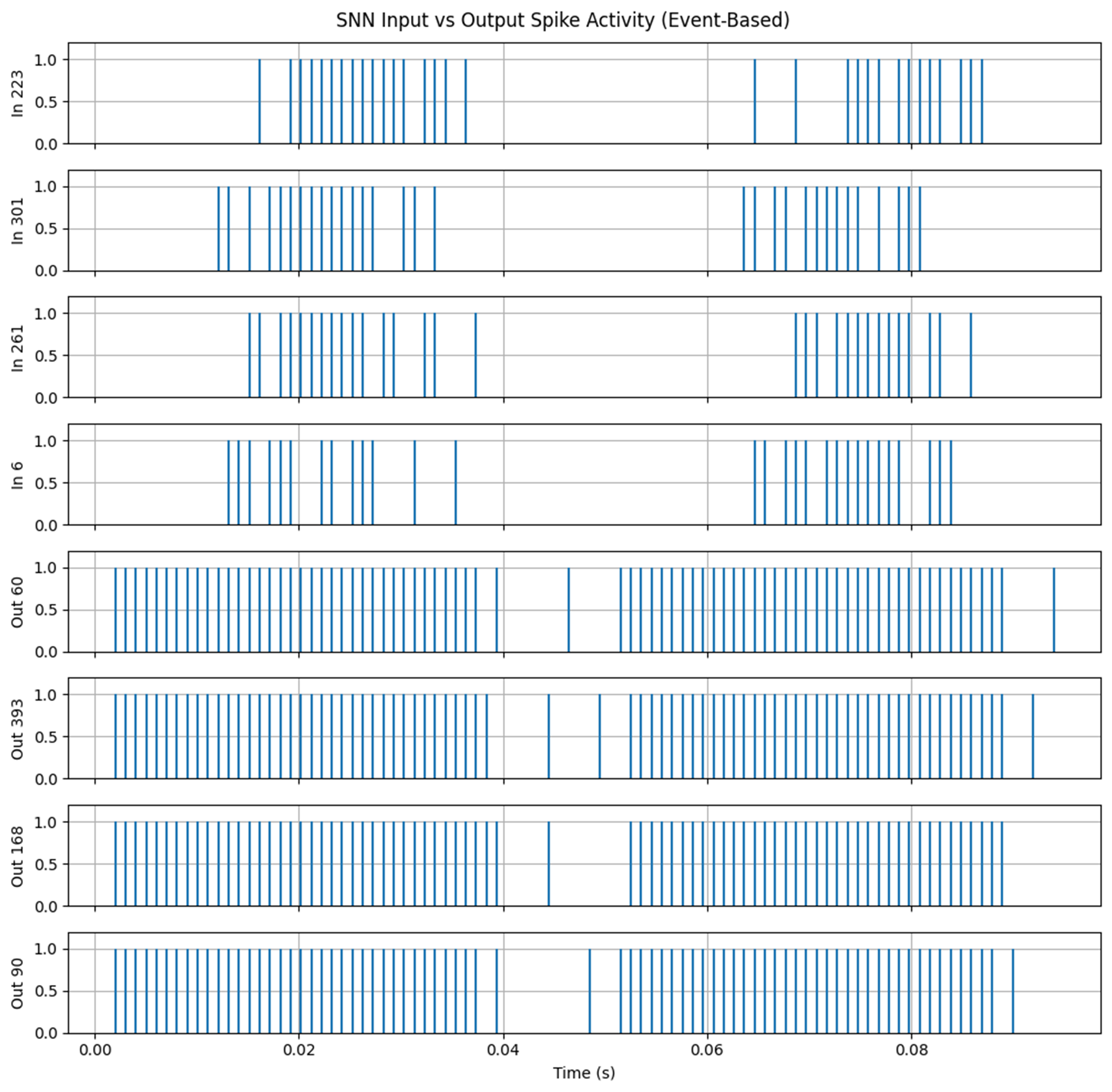}
    \caption{Input and output spike activity for selected channels in the SNN model.}
    \label{fig:snn_spikes}
\end{figure}

For comparison, Figure~\ref{fig:snn_spikes} shows the corresponding input and output spike activity of a spiking neural network (SNN) trained on the same data.

As shown in Figure~\ref{fig:snn_spikes}, the input consists of sparse and irregular spike events distributed over time, similar to those presented in Figure~\ref{fig:input_spikes}. The output of the SNN exhibits a transformed spike pattern, where neurons respond with sequences of discrete firing events. In several cases, temporally clustered spikes can be observed, indicating that the network integrates input over short time windows and produces burst-like activations.

Compared to the continuous wave-based representation in Figure~\ref{fig:wave_pulses}, the SNN maintains a strictly discrete event-based encoding. While temporal structure is present in the form of spike timing and clustering, the representation remains binary and lacks an explicit notion of continuous activation magnitude.

This highlights a fundamental difference between the two approaches: the SNN represents information through the occurrence and timing of discrete events, whereas the proposed wave-based model encodes information in continuous energy distributions over time. As a result, the wave-based representation provides a smoother and more continuous description of temporal structure, while the SNN retains a sparse and event-driven formulation.

To evaluate the effectiveness of the proposed Continuous Temporal Energy Network (CTEN), we compare its performance against a spiking neural network (SNN) baseline on the interaural phase difference (IPD) classification task. 

Table~\ref{tab:ipd_comparison} summarizes the observed accuracy ranges across multiple training runs, together with the corresponding training times measured on an NVIDIA T4 GPU. The comparison highlights differences in predictive performance, computational efficiency, and representation type between the two approaches.

\begin{table}[width=\linewidth,cols=4,pos=h]
\caption{Comparison of model performance on the IPD classification task. Accuracy is reported as the observed range over multiple runs. Training time is measured on an NVIDIA T4 GPU (Google Colab environment).}
\label{tab:ipd_comparison}

\begin{tabular*}{\tblwidth}{@{}lccc@{}}
\toprule

\textbf{Model} &
\textbf{Accuracy (\%)} &
\textbf{Training Time (s)} &
\textbf{Representation} \\

\midrule

CTEN (proposed)
& 90 -- 94
& 6
& Continuous (wave-based) \\

SNN
& 80 -- 94
& 36
& Discrete (event-based) \\

\bottomrule
\end{tabular*}
\end{table}

The results indicate that the proposed CTEN model achieves stable and consistent performance across runs, while maintaining competitive accuracy with significantly reduced training time. In contrast, the SNN baseline exhibits higher variability and increased computational cost.
\subsection{Preliminary sEMG Validation on Ninapro}

The purpose of this experiment is not to establish state-of-the-art performance, but to evaluate whether the proposed continuous temporal energy representation generalizes beyond synthetic benchmarks to real-world multichannel biosignals.

The proposed CTEN model was evaluated on the Ninapro DB1 dataset using raw temporal sEMG windows with minimal preprocessing and without handcrafted feature engineering. To ensure a controlled comparison, all models were trained under approximately matched parameter budgets and evaluated across multiple random seeds.

Table~\ref{tab:ninapro_results} summarizes the experimental configuration and performance statistics of the proposed CTEN model on the Ninapro DB1 sEMG dataset. The evaluation was conducted using raw multichannel temporal sEMG windows with minimal preprocessing and without handcrafted feature extraction. Results are reported over twenty independent random seeds in order to assess both predictive performance and training stability.

\begin{table}[width=\linewidth,cols=2,pos=h]
\caption{Performance of the proposed CTEN model on the Ninapro DB1 sEMG dataset using raw multichannel temporal signals. Results are reported over 20 random seeds.}
\label{tab:ninapro_results}

\begin{tabular*}{\tblwidth}{@{}lc@{}}
\toprule

\textbf{Metric} &
\textbf{Value} \\

\midrule

Dataset &
Ninapro DB1 \\

Input channels &
10 \\

Window length &
100 samples \\

Number of classes &
17 \\

Training epochs &
40 \\

Model type &
CTEN \\

Hidden dimension ($H$) &
280 \\

Interaction rank ($R$) &
64 \\

Parameters &
120476 \\

Mean accuracy (\%) &
80.10 \\

Accuracy std &
1.12 \\

Best accuracy (\%) &
82.66 \\

Worst accuracy (\%) &
78.63 \\

Mean training time (s) &
3.44 \\

Preprocessing &
Normalization only \\

Feature engineering &
None \\

Input representation &
Raw temporal sEMG \\

\bottomrule
\end{tabular*}
\end{table}

For comparison, Table~\ref{tab:ninapro_compare} presents the performance of CTEN alongside parameter-matched CNN and LSTM baselines trained on the same temporal input representation.
\begin{table}[width=\linewidth,cols=5,pos=h]
\caption{Parameter-matched comparison on the Ninapro DB1 sEMG dataset using raw temporal input signals. Results are reported as mean accuracy over 20 random seeds.}
\label{tab:ninapro_compare}

\begin{tabular*}{\tblwidth}{@{}lcccc@{}}
\toprule

\textbf{Model} &
\textbf{Parameters} &
\textbf{Mean Acc. (\%)} &
\textbf{Std} &
\textbf{Representation} \\

\midrule

CTEN (proposed)
& 120476
& 80.10
& 1.12
& Continuous temporal energy \\

1D CNN
& 124076
& 78.57
& 3.49
& Convolutional temporal features \\

LSTM
& 140076
& 61.17
& 4.86
& Recurrent temporal dynamics \\

\bottomrule
\end{tabular*}

\end{table}
The results indicate that the proposed CTEN achieves competitive performance under approximately matched parameter budgets while exhibiting substantially lower variance across random seeds compared to the CNN baseline.

Interestingly, the CTEN model achieved both the highest average accuracy and the lowest observed variability, suggesting that the continuous temporal energy representation provides stable convergence behavior for short-window event-driven biosignals.

In contrast, the recurrent LSTM baseline demonstrated significantly lower performance and considerably higher variability across runs. This observation may indicate that the considered short-window temporal classification task benefits more from localized temporal aggregation than from long-range recurrent state propagation.

Overall, these findings suggest that the proposed continuous temporal energy representation captures relevant temporal structure directly from raw sEMG signals without relying on explicit recurrent dynamics or handcrafted temporal feature extraction.
\subsection{CTEN-TA: Temporal Attention Extension}

To further improve the ability of CTEN to exploit temporal dependencies in multichannel sEMG signals, a temporal attention mechanism was introduced, resulting in the proposed \textbf{CTEN-TA} (Complex Temporal Encoding Network with Temporal Attention) architecture. The original CTEN computes a complex-valued phase representation followed by low-rank interaction modeling and energy estimation. In CTEN-TA, a single multi-head self-attention block is inserted after the power representation stage and before global feature aggregation.

Given the power representation
\[
P \in \mathbb{R}^{T \times H},
\]
queries, keys, and values are computed through learned linear projections

\[
Q = PW_Q,
\qquad
K = PW_K,
\qquad
V = PW_V,
\]

where $W_Q$, $W_K$, and $W_V$ are trainable projection matrices. Temporal self-attention is then computed as

\[
A =
\mathrm{softmax}
\left(
\frac{QK^\top}{\sqrt{d}}
\right)V,
\]

allowing the network to identify and reinforce recurring activation patterns occurring across multiple EMG pulses within a temporal window. A residual connection and layer normalization are subsequently applied,

\[
P_1 =
\mathrm{LayerNorm}
\left(
P + A
\right),
\]

followed by a feed-forward network (FFN),

\[
F
=
W_2\,
\mathrm{GELU}
\left(
W_1 P_1
\right),
\]

and a second residual normalization stage,

\[
P_2
=
\mathrm{LayerNorm}
\left(
P_1 + F
\right).
\]

Finally, global mean pooling and max pooling are concatenated and passed to the classifier. The resulting architecture preserves the original CTEN phase--interaction framework while enabling explicit modeling of temporal relationships between EMG activations.

Experimental evaluation revealed that a single attention block consistently improved performance, whereas deeper attention stacks and attention-based pooling reduced classification accuracy. The best results were obtained using eight attention heads, indicating that multiple parallel attention mechanisms can capture complementary temporal relationships without excessive model complexity.

\subsection{Ablation Study and Comparison with CNN-BiLSTM}

Table~\ref{tab:cten_ablation} summarizes the performance of the proposed CTEN variants and compares them against a CNN-BiLSTM baseline on the Ninapro DB1 dataset.

\begin{table}[ht]
\centering
\caption{Ablation study of CTEN variants and comparison with CNN-BiLSTM on Ninapro DB1. Results are reported as mean accuracy over multiple random seeds.}
\label{tab:cten_ablation}
\begin{tabular}{lrrrr}
\toprule
\textbf{Model} &
\textbf{Parameters} &
\textbf{Mean Acc. (\%)} &
\textbf{Std (\%)} &
\textbf{Best (\%)} \\
\midrule

CTEN &
55,767 &
83.86 &
-- &
83.86 \\

CTEN-TA (4 heads) &
328,715 &
85.18 &
0.89 &
86.37 \\

CTEN-TA (8 heads) &
\textbf{178,071} &
\textbf{87.53} &
\textbf{0.52} &
\textbf{88.05} \\

CTEN-TA + Attention Pooling &
161,815 &
78.20 &
0.21 &
78.41 \\

CTEN-TA2 (2 Attention Blocks) &
310,551 &
82.91 &
0.31 &
83.23 \\

CNN-BiLSTM &
344,791 &
82.81 &
1.05 &
83.86 \\

\bottomrule
\end{tabular}
\end{table}

The ablation study demonstrates that the introduction of a single temporal attention block substantially improves the original CTEN architecture. Increasing the number of attention heads from four to eight further improved performance, yielding the highest observed mean accuracy of $87.53\%$. In contrast, attention pooling and deeper attention stacks reduced performance, suggesting that discriminative information is distributed across the temporal window and is best captured through a single attention layer followed by global mean--max feature aggregation.

Notably, the proposed CTEN-TA achieved higher accuracy than the CNN-BiLSTM baseline while using approximately half the number of trainable parameters. These results indicate that temporal attention complements the phase-based representation of CTEN and provides an efficient mechanism for modeling recurring EMG activation patterns.
\section{Discussion}

The proposed framework demonstrates that asynchronous event-driven activity can be effectively represented in a continuous temporal domain through energy-based latent representations. By transforming sparse temporal events into continuous activation fields, the proposed Continuous Temporal Energy Network (CTEN) enables fully differentiable optimization while preserving relevant temporal structure.

The experimental results indicate that the proposed model achieves stable and competitive performance across both synthetic and real-world temporal benchmarks. On the synthetic interaural phase difference (IPD) task, CTEN achieved classification performance comparable to spike-based neural models while maintaining substantially reduced training time and stable convergence behavior across random seeds.

More importantly, the experiments on the Ninapro DB1 sEMG dataset demonstrate that the proposed representation generalizes beyond synthetic event streams to real multichannel biosignals. Under approximately matched parameter budgets, CTEN achieved higher average accuracy and lower variance than both convolutional and recurrent baselines.

An additional observation emerged when extending CTEN with a temporal self-attention mechanism, resulting in the proposed CTEN-TA architecture. While the original CTEN already demonstrated competitive performance on the Ninapro DB1 benchmark, the introduction of a single multi-head temporal attention block consistently improved classification accuracy across multiple random seeds. The best configuration was obtained using eight attention heads, yielding a mean accuracy of 87.53\% while requiring substantially fewer parameters than the CNN-BiLSTM baseline.

Interestingly, further increases in architectural complexity did not lead to additional improvements. Experiments involving deeper attention stacks and attention-based pooling mechanisms resulted in reduced performance, suggesting that the discriminative information is distributed across the temporal window rather than concentrated at a small number of highly informative time points. This observation is consistent with the hypothesis that repeated and correlated EMG activation patterns contribute collectively to the classification decision.

The ablation study therefore indicates that the primary benefit originates from the introduction of a single temporal attention stage capable of reinforcing recurring activation structures within the latent energy representation. Additional attention layers appear to introduce unnecessary complexity for the considered short-window sEMG classification task. These findings suggest that CTEN-TA achieves an effective balance between representational capacity and model simplicity, providing both improved accuracy and reduced sensitivity to random initialization compared with conventional CNN-BiLSTM architectures.

In particular, the comparison with the parameter-matched CNN-BiLSTM baseline is notable. Despite using a comparable number of trainable parameters, CTEN-TA achieved both higher average accuracy and lower variance across random seeds. The reduced sensitivity to initialization suggests that the combination of phase-based latent representations and temporal attention provides a robust mechanism for modeling short-window multichannel biosignals.

The recurrent LSTM baseline demonstrated substantially lower performance and significantly higher variability despite having a comparable parameter count. This observation may indicate that the considered short-window temporal classification task benefits more from localized temporal aggregation than from long-range recurrent state propagation.

From a representational perspective, the proposed framework differs fundamentally from conventional spike-based formulations. The SNN baseline encodes information through sparse discrete firing events, whereas CTEN represents temporal structure through continuous energy distributions evolving over finite observation windows. Although both approaches operate on temporally structured data, the resulting latent representations exhibit markedly different characteristics.

The visualization of latent activations further illustrates this distinction. CTEN produces structured wave-like activation patterns with localized temporal energy concentrations, whereas the SNN maintains sparse binary spike responses. This suggests that temporally structured event information can be represented either through discrete event timing or through continuous temporal energy fields.

Importantly, the present results do not suggest that continuous temporal representations replace spike-based models in general. Instead, the findings indicate that continuous temporal energy representations may provide a computationally efficient and stable alternative for certain classes of event-driven temporal learning problems, particularly when the observable signal already corresponds to aggregated meso-scale activity, such as sEMG recordings.

The proposed model can also be interpreted as a learned continuous alternative to kernel-based temporal smoothing, where the temporal representation is optimized directly through gradient-based learning rather than predefined convolution kernels.

While the present study demonstrates promising behavior on controlled synthetic data and preliminary sEMG benchmarks, several limitations remain. First, the current formulation does not explicitly model discrete spike generation or biologically realistic neuronal dynamics. Consequently, the framework should primarily be interpreted as a continuous temporal representation model rather than a biologically faithful neural simulator.

Second, the present experiments focus primarily on short-window temporal classification tasks. Additional studies on larger-scale temporal datasets and more complex event-driven learning problems are required to fully characterize the representational properties and scalability of the proposed framework.

Future work will therefore focus on extending the evaluation to additional biosignal datasets, investigating alternative temporal aggregation mechanisms, and further analyzing the relationship between continuous temporal energy representations and conventional event-driven neural computation.

\section{Conclusion}

In this work, we introduced the Continuous Temporal Energy Network (CTEN), a continuous temporal modeling framework for asynchronous event-driven signals based on phase-modulated latent energy representations. The proposed approach transforms sparse temporal activity into structured continuous energy patterns over finite observation windows, enabling fully differentiable learning without explicit recurrence, discrete spike simulation, or surrogate gradient approximations.

Experimental evaluation on a synthetic interaural phase difference (IPD) classification task demonstrated that the proposed framework achieves stable and competitive performance while maintaining efficient training dynamics. Furthermore, experiments on the Ninapro DB1 sEMG benchmark showed that the proposed representation generalizes beyond synthetic event streams to real-world multichannel biosignals using raw temporal input signals with minimal preprocessing.

To further exploit temporal dependencies, a temporal self-attention extension, denoted CTEN-TA, was introduced. The addition of a single multi-head attention block consistently improved classification performance, while deeper attention stacks and attention-based pooling mechanisms did not provide additional benefits. The best-performing configuration employed eight attention heads and achieved a mean classification accuracy of 87.53

The conducted ablation studies indicate that the combination of continuous temporal energy representations and a lightweight temporal attention mechanism provides an effective balance between representational capacity, parameter efficiency, and training stability. The results further suggest that recurring activation structures within multichannel biosignals can be modeled effectively through temporal attention operating on latent energy representations.

Overall, the proposed framework provides a complementary perspective to conventional spike-based, convolutional, and recurrent temporal models by representing event-driven activity through continuous latent energy dynamics that remain computationally efficient and fully compatible with standard gradient-based optimization.

Future work will focus on extending the evaluation to additional biosignal datasets, investigating the role of temporal attention in latent phase representations, and exploring the applicability of continuous temporal energy modeling to broader classes of event-driven sensing and biomechanical control problems.

\section*{Acknowledgments}
 We acknowledge NVIDIA Corporation for supporting this work with GPU hardware through a donated RTX 6000 unit.


\bibliographystyle{cas-model2-names}

\newpage
\bibliography{references_prob2026}

\newpage
\appendix
\section{Reduced Two-Channel Wave Model}

To provide an interpretable illustration of the proposed framework, we consider a reduced configuration with two input channels and four latent units. This simplified setting allows explicit visualization of how discrete spike inputs are transformed into continuous wave-based representations.

\subsection{Input Configuration}

We consider an input signal of the form $x(t) \in \mathbb{R}^{2}$, where each component corresponds to a spike train:
\[
x(t) =
\begin{bmatrix}
x_1(t) \\
x_2(t)
\end{bmatrix}.
\]

The two channels represent temporally shifted pulse trains, where channel 1 activates earlier in time and channel 2 exhibits a delayed activation pattern.

\subsection{Linear Projection}

The input is mapped to a latent space of dimension $H = 4$:
$h(t) = W^\top x(t), \quad W \in \mathbb{R}^{2 \times 4}.$

Each latent unit therefore represents a weighted combination of the two input channels.

\subsection{Temporal Integration}

To illustrate temporal smoothing, we introduce a simple exponential accumulation:
\[
\tilde{h}(t) = \sum_{\tau \le t} e^{-\lambda (t-\tau)}\, h(\tau)
\]
where $\lambda > 0$ controls the decay rate.

This step highlights how discrete spike events induce continuous temporal responses.

\subsection{Wave Embedding}

Each latent unit is modulated by a phase function:
\[
\theta_h(t) = \omega_h t + \phi_h
\]

leading to the real-valued wave representation:
\[
\psi_h(t) = \tilde{h}_h(t)\cos\!\big(\theta_h(t)\big)
\]
This formulation encodes temporal structure through oscillatory behavior rather than explicit recurrence.

\subsection{Energy Representation}

The corresponding energy signal is defined as
\[
P_h(t) = \psi_h(t)^2
\]

This transformation removes phase dependence while preserving temporal localization and amplitude information.

\subsection{Interpretation}

\begin{figure}[h]
\centering
\includegraphics[width=\linewidth]{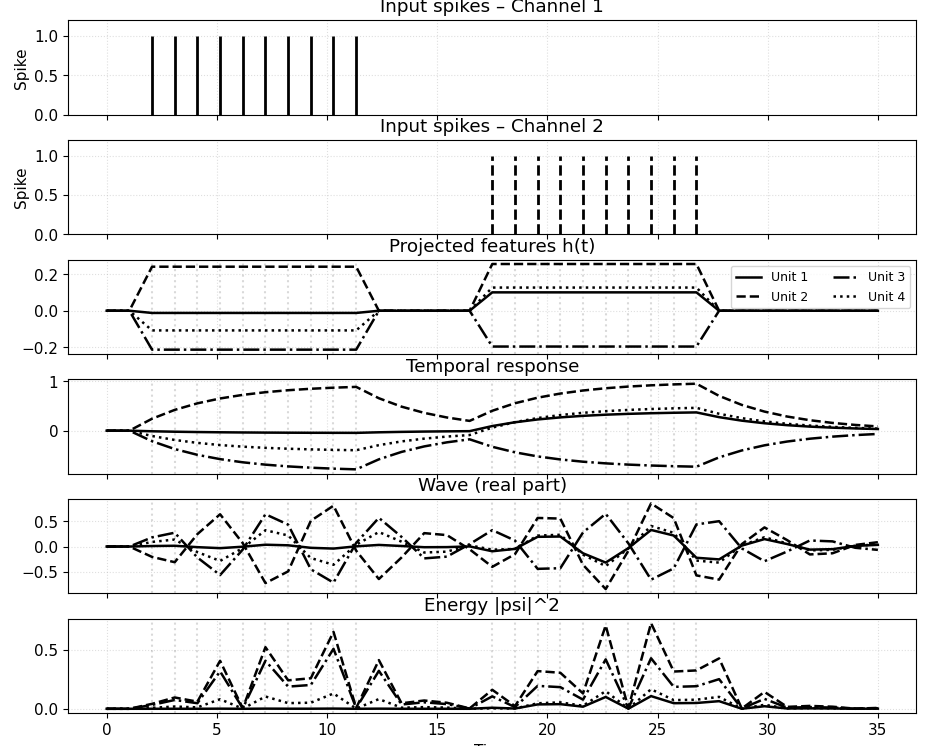}
\caption{Illustration of the reduced two-channel wave model.}
\label{fig:appendix_reduced}
\end{figure}

Figure~\ref{fig:appendix_reduced} illustrates the transformation pipeline:

\begin{itemize}
\item Two temporally shifted spike trains (input channels),
\item Linear projection into four latent units,
\item Temporal integration producing smooth activation profiles,
\item Wave modulation introducing oscillatory structure,
\item Energy projection yielding localized activation patterns.
\end{itemize}

The reduced model demonstrates how the proposed framework converts sparse, discrete spike events into structured continuous representations.

In particular, the temporal offset between the two input channels is preserved and becomes visible in both the latent activations and the resulting energy distributions.

Notably, even in this reduced setting, the transformation is nonlinear due to the combination of temporal integration and phase modulation.

\end{document}